# gprHOG and the popularity of Histogram of Oriented Gradients (HOG) for Buried Threat Detection in Ground-Penetrating Radar

Daniel Reichman, Leslie M. Collins, *Senior Member, IEEE*, and Jordan M. Malof, *Member, IEEE*

*Abstract*— Substantial research has been devoted to the development of algorithms that automate buried threat detection (BTD) with ground penetrating radar (GPR) data, resulting in a large number of proposed algorithms. One popular algorithm GPR-based BTD, originally applied by Torrione et al., 2012, is the Histogram of Oriented Gradients (HOG) feature. In a recent large-scale comparison among five veteran institutions, a modified version of HOG referred to here as "gprHOG", performed poorly compared to other modern algorithms. In this paper, we provide experimental evidence demonstrating that the modifications to HOG that comprise gprHOG result in a substantially better-performing algorithm. The results here, in conjunction with the large-scale algorithm comparison, suggest that HOG is not competitive with modern GPR-based BTD algorithms. Given HOG's popularity, these results raise some questions about many existing studies, and suggest gprHOG (and especially HOG) should be employed with caution in future studies.

## I. Introduction

In this work we focus on the problem of developing algorithms for automatic buried threat detection (BTD) using 2-dimensional measurements from a ground penetrating radar (GPR), called B-scans [1]–[3]. Figure 1 presents examples of B-scans collected over the locations of buried threats. This topic has received substantial attention in recent years, resulting in a large number of proposed algorithms [4]–[7].

Most recently proposed algorithms have adopted some variant of a general processing pipeline comprised of two steps: *feature extraction* and *classification*. In feature extraction, a function is designed that extracts a numeric vector representing the GPR imagery. Ideally, the extracted vector will succinctly encode any visual content that is relevant to decision-making, while suppressing irrelevant content such as noise. The classification step is comprised of a function that maps the feature vectors to a decision statistic, or confidence, indicating the relative likelihood that the vector corresponds to a buried threat.

In 2014, Torrione et al. [5] proposed to use the Histogram of Oriented Gradients (HOG) feature from the computer vision community for GPR-based BTD, and it has subsequently become widely used [8]–[19] (88 total Google Scholar citations, at time of writing). In particular, HOG is frequently used as a baseline approach to suggest the superiority of new algorithms [8], [9], [13]–[18], [20]–[22]. Surprisingly however, a modified version of HOG, called "gprHOG", recently performed poorly in a large-scale algorithm comparison, conducted with algorithms submitted by several institutions with substantial experience in GPR-based BTD[7]. The results in [7] however can justifiably be viewed with skepticism because it provides relatively little detail about gprHOG, and no experimental evidence justifying its superiority over HOG.

In this work we use a large collection of real-world GPR data, and two popular experimental designs, to show that gprHOG substantially and consistently outperforms HOG. In conjunction with the findings in [7], this strongly suggests that HOG is not competitive with other modern algorithms. Given the widespread popularity of HOG, this is an important finding and it suggests that gprHOG (and especially HOG) should be employed with caution in future studies.

Despite the relatively poor performance of gprHOG in [7], we show that each of its modifications to HOG yield (sometimes large) performance improvements. These modifications are largely generic, and could be used to improve other GPR-based algorithms, or to guidance for new algorithm designs.

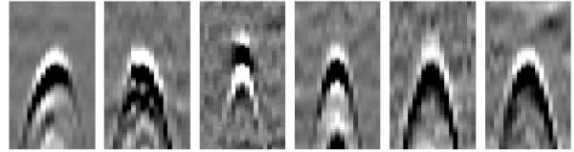

Figure 1: Examples of GPR data collected over several different buried threats. In each panel, referred to as a B-scan, the vertical axis corresponds to time (or sometimes depth), and the horizontal axis corresponds to space.

The remainder of this paper is organized as follows: Section II presents the GPR system and data; Section III presents background information; Section IV presents the experimental design; Section V presents the results of each modification to HOG, and the final gprHOG algorithm; and Section VI presents conclusions.

## II. GPR System and Dataset Description

All data in this work were collected using a downward looking GPR (similar to the one in [1], [6], [7]). The GPR is comprised of an array of equally spaced antennas (i.e., cross-track) that are pointed perpendicularly downward so that data can be collected at subsurface locations directly below each individual antenna. Each antenna emits an ultra-wideband electromagnetic signal, consisting of a differentiated Gaussian pulse, and then measures the energy reflected back from the subsurface. Each antenna collects a time-series of reflection energies, known as an A-scan [1]. As the vehicle travels down a lane, A-scans are collected at regular spatial intervals (i.e., downtrack) from each antenna. The measured A-scans result in a 3-D cube of GPR data with two spatial dimensions and one temporal dimension, as shown in Figure 2(a).

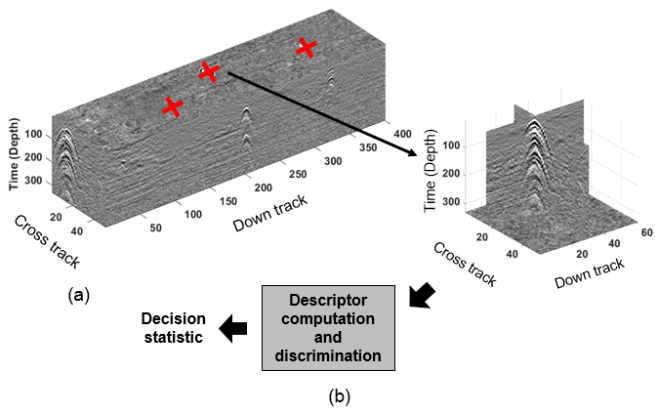

**Figure 2:** Illustration of a typical processing pipeline in the GPR BTD literature. The goal is to process the entire cube of data and provide a probability of buried threat presence at each location in the cube. (a) The GPR volume is processed by a computationally inexpensive algorithm, referred to as a prescreener, to identify suspicious locations, which are indicated by red crosses. (b) At each suspicious location, a feature vector is computed using surrounding GPR data. The feature vector is provided to a discrimination algorithm to compute the likelihood of threat presence at this location. The output is a scalar decision statistic, or "confidence".

The GPR system was used to collect data at two U.S. test sites, comprising a total surface area of $120,817.4$ m$^2$. The data was collected over 13 distinct lanes especially prepared for this purpose. A total of 90 runs were made over all of the lanes. In each lane, targets were buried at known locations, to facilitate evaluation of the developed BTD systems. A total of 664 unique targets were buried for this work and are of varied types and metal content and were buried at a range of depths of interest. The total dataset consists of 4,552 threat encounters.

### III. BACKGROUND METHODS

In this section, we provide background information regarding the methods used in this work. This includes a description of the framework we follow for designing and evaluating BTDs and a description of the HOG feature.

#### A. The detection pipeline

The cube of collected GPR data is processed in a pipeline that is typically used for BTDs in this literature [1], [5]–[7], [23] and is illustrated in Figure 2. The processing pipeline is divided into two stages: prescreening and discrimination. The prescreener flags spatial locations in the cube that it considers anomalous or otherwise suspicious. The discriminator processes the data at those locations, referred to as alarm locations, and outputs a decision statistic for each, indicating the relative likelihood of buried threat at that location. The focus of this work is on discrimination algorithms (i.e., feature extraction plus classification).

#### B. Data extraction at prescreener alarm locations

The discriminator is used to further inspect the data at locations flagged by the prescreener. A large portion of the data at those locations corresponds to background data; the reflections that would commonly be associated with threat presence are typically within a relatively small and localized spatio-temporal extent [24]. The discriminator is typically provided with data from a subset of the larger cube that encompasses this extent for each alarm. This subset cube is extracted around spatio-temporal coordinates of interest in the volume, referred to as keypoints [1]. Thus, at each threat and non-threat spatial location flagged by the prescreener, temporal coordinates where buried threat signal exists at a particular spatial location, need to be identified. The Maximum Smoothed Energy Keypoint (MSEK) method is used here for this purpose, as described in [25]. This method operates in several steps. The initial ground reflection is first estimated, aligned, and then removed (see Section IV.A). Subsequently, the B-scan is background normalized, its central A-scan smoothed and local maxima are chosen as candidate locations for sub-image extraction.

#### C. The HOG Feature

The HOG feature was designed to succinctly encode shape information of natural imagery [26]. In the context of GPR data, HOG is used to encode the hyperbolic shape that buried threat signals typically exhibit [5]. The HOG feature is computed in three steps, which are described in detail in Figure 3, and are computed on the image patches extracted at keypoints in the GPR volume. The output from this step is a vector of numbers, which are provided to the discriminator in place of the raw pixel intensities of the image patch.

#### D. Supervised classification

To perform discrimination between the threat and non-threat classes, we follow the common approach of applying supervised classifiers [1], [7], [23]. Supervised classifiers perform discrimination by first learning the difference between the two classes from labeled examples provided to the classifier in a process referred to as training. Once trained, the classifier can be applied to new data to determine whether a threat is buried at that new location. To train the detector, data is used that is extracted at prescreener alarm locations for threats and non-threats.

#### E. Evaluating algorithm performance

In this work, we trained and tested each discriminator using cross-validation. This is a common approach for evaluating the performance of machine learning algorithms, and has been employed previously for BTD with GPR data [1], [5], [6], [16]. In this procedure, the dataset is divided into $N$ non-overlapping groups, referred to as folds. The discriminator is trained on $N-1$ folds and the resulting model is used to predict threat presence on the outstanding fold. This process is repeated $N$ times in cyclical fashion where each time, a model is trained to obtain predictions on one of the folds that have not yet been used for testing. In this way, predictions are made on the entire dataset without ever using a discriminator for testing that was also trained on that same fold.

To compare the detection performance of each trained classifier, receiver operating characteristic (ROC) curves are commonly used in the BTD algorithm research literature [6], [27]. ROC curves plot the relationship between the false detection rate (x-axis) and true detection rate (y-axis) of a classifier, as its sensitivity is varied. We adopt the common practice in the BTD literature of scaling the x-axis of the ROC curve to report the false alarm rate in terms of false alarms per square meter [6], [23]. In this work, the exact FAR range is redacted. The ROC is constructed using prescreener alarms where an alarm is declared a true threat if it lies within 0.25m of a threat ground truth location. The discriminator decision statistic is used for each alarm.

## IV. EXPERIMENTAL DETAILS

In this section, we provide details regarding specific design choices made for the experiments in this work. The design choices are kept as similar as possible to the original application of HOG by Torrione et al., 2014, [5] to facilitate comparison.

### A. Prescreening and data processing

The dataset for algorithm development is obtained by using the combination prescreeners described in [7]: F2, an energy-based anomaly prescreener and CCY, a shape-based prescreener. Alarms are obtained by setting a sensitivity threshold for the prescreener such that 96.05% of threat locations are identified corresponding to a total 4,372 threat locations along with 6,070 non-threat locations [7].

At each prescreener alarm location, several common preprocessing steps are applied to the cube of data before feature computation [1], [5], [6]. The initial cube of data consists of 448 time indexes. First, the time index at which the ground reflection first appears (i.e., the boundary of the air-ground interface) is estimated. Each A-scan is then shifted up or down to align the data so that the ground appears at the same time index of 100. The response in air and slightly below the aligned ground is subsequently removed and the data from times 109 to 448 are kept for further processing. The resulting cube of 330 time samples is subsequently depth normalized.

### B. HOG feature parameters

The parameters used for the HOG feature are slightly modified to better capture the average shape of threats in our dataset and to avoid computing the feature on much background data. An image size of $18 \times 20$ pixels (time $\times$ space coordinates) is used and following [5], we use a $3 \times 4$ cell configuration. Therefore, we use cells of size $6 \times 5$ pixels. Both the block size and the number of angle bins are kept the same: blocks are composed of $3 \times 3$ cells and the histogram is constructed with $\theta = 9$ angle bins.

### C. Classification

For classification, we use the same classifier as [5], namely the Random Forest classifier (MATLAB implementation using 100 trees, 2 variable splits at nodes, and leaf-size of 1) [28]. For each alarm, two HOG feature vectors are computed using the same parameters: one on an image extracted with A-scans along the cross-track (cross-track B-scan) and the other on an image with A-scans along the down-track direction (down-track B-scan). The two feature vectors are concatenated to form a single vector, which is provided to the classifier.

Because multiple sub-images in a B-scan can be relevant for detection and this number may differ between threats and non-threats, the training dataset construction is an additional design parameter. In [5], for threats, MSEK is used to identify the top 4 locations with highest signal energy. For non-threats, sub-images are selected at regular, small intervals down the temporal axis of the B-scan. When applying the BTD to data at a new spatial location, the trained classifier is applied to every $4^{th}$ temporal location at that spatial location, resulting in 82 total possible testing locations. The final confidence for scoring is computed as the sum of the confidences at the $L$ temporal locations with highest confidence, where $L = 3$ in [5].

### D. Cross-validation (OBCV and LBCV)

We use two types of cross-validation to account for the possible change in performance for different such procedures [29]. The first type is lane-based cross-validation (LBCV) where each fold corresponds to the data of a particular lane, resulting in 13 folds. The second type is referred to as object-based cross-validation, OBCV, (or stratified cross-validation) [1], [5], [6]. In this scenario, individual objects on the lane are assigned to folds, 10 in this case, independent of the lane they come from. We additionally note that the HOG feature was developed for OBCV and therefore we want to measure whether performance is improved in that scenario as well.

## V. THE GPRHOG ALGORITHM

In this section, we present experiments that were used to investigate the effect the adaptations to the computation of the HOG feature we suggest have on discrimination performance. The adaptations we propose here are aimed at tailoring the computation of HOG to GPR data (hence the name gprHOG) and are therefore expected to be beneficial across datasets. We note that we have found these changes to be beneficial across several datasets and radars, but those results are not shown in this paper. The experimental design described in Section IV are used to evaluate the performance of the proposed adaptations.

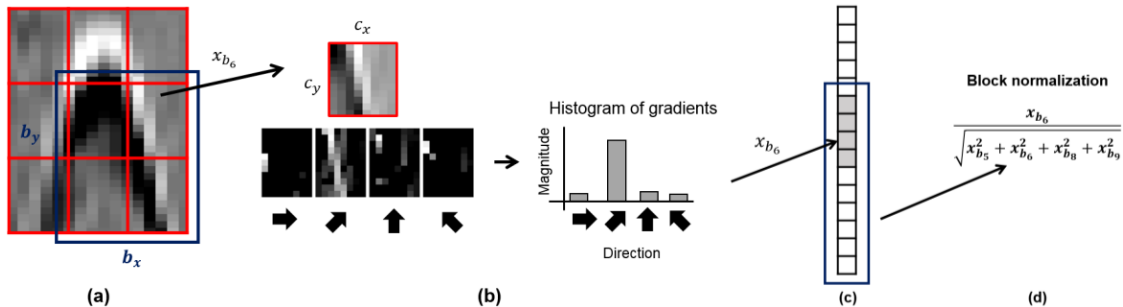

Figure 3: Illustration of the HOG feature computation. (a) The image patch is divided into a grid of sub-windows, referred to as cells, of dimension $c_x \times c_y$ pixels, which are demarcated by red boxes. The green box demarcates a block of $b_x \times b_y$ cells, which will be used in a later step. (b) In each cell, the direction of the gradient at each pixel is computed. A histogram of the directions of the pixels in the cell is computed with $\theta$ angle bins whose centers are evenly distributed between 0 and 180 degrees. Each pixel is assigned to the angle bin to which its gradient direction is most similar. The contribution of each pixel to the histogram is weighted by the gradient magnitude of that pixel. An example of the computation of the histogram is shown for the sixth block, $x_{b_6}$, with $\theta = 4$. (c) In this fashion, a histogram is computed for each cell and the histogram values of each cell are organized as a vector of numbers. (d) The histograms are normalized using the block of neighboring cells to the current cell (including the current cell, as shown in the green box in (a)). In this example, the histogram entries of $x_{b_6}$ are divided by the $l_2$ norm of the entries of the histograms of the cells in the normalization block.

*A. Removal of block normalization*

The first adaptation regards the last step in the HOG computation process, namely, the block normalization. This step is used in natural imagery to mitigate local contrast changes [26]. In the context of GPR, regions of high contrast are typically indicative of the presence of a subsurface anomaly (e.g., a buried threat). Therefore, in this experiment, we remove the normalization step altogether. The performance difference shown in Figure 4 (blue lines versus red lines) suggests that block normalization does not translate well to GPR data.

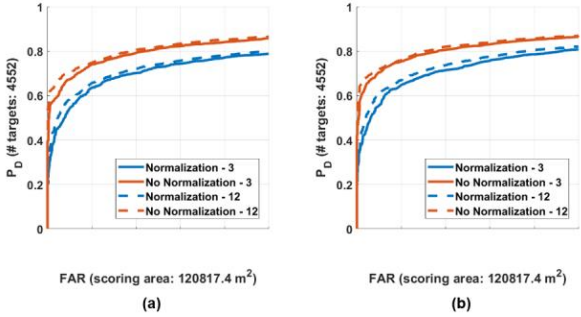

**Figure 4:** The results of removing the block normalization step from the HOG feature, increasing $L = 3$ to $L = 12$ similar to the conclusions from [1]. Result when evaluating the algorithm using **(a)** LBCV, **(b)** OBCV.

*B. Greater temporal keypoint averaging*

Recent work in [1] demonstrated that, when using shorter patches such as those employed by HOG (in contrast to e.g., EHD, SED, and LG feature [7]), it is generally best to average classifier predictions over a relatively larger number of temporal keypoints, $L$, along the central A-scan of a prescreener alarm. This can be done to account for the possibly diffuse reflections over time, or multiple reflections buried threats often exhibit.

In [1], the authors found that the best-performing value of $L$ will vary across classifiers and datasets, but that performance varies little when $L \in [8,12]$. Furthermore, performance within this range is always superior to the value of $L = 3$ typically used with HOG. In our experiments $L = 12$ similarly yielded the best performance, as shown in Figure 4, but similar to [1], performance varied little with respect to $L$.

*C. Multiple B-scan feature averaging*

The third adaptation we propose is motivated by the observation that GPR data is relatively noisy compared to natural imagery. Furthermore, the patterns we wish to identify in GPR-based BTD are relatively simple. In such a context, an approach to mitigate noise is to average raw GPR data, or feature vectors, where possible.

To compute the gprHOG feature, for a given direction (e.g., down-track), 3 additional down-track HOG features should be extracted on B-scans on either side of the prescreener alarm location (i.e., 7 HOG feature vectors in total). A final down-track HOG feature is computed by averaging these 7 vectors together. The same procedure is performed for the cross-track direction. In Figure 5, we present the results of this feature averaging approach, where the averaged HOG features are referred to as "denoised HOG." We experimented with averaging different numbers of B-scan features, and while they all improved performance, we found that 3 on either side performed best. That averaging the feature vector improves performance is consistent with our recent work in [30], as well.

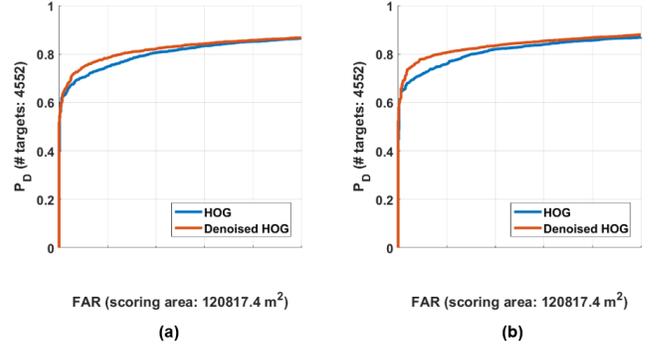

**Figure 5:** Classification performance using the proposed denoising post-processing step to the HOG feature. Performance is evaluated using **(a)** LBCV, **(b)** OBCV.

*D. gprHOG versus the original HOG*

Figure 6 presents a comparison of gprHOG and HOG (from Torrione et. al., 2014). gprHOG substantially outperforms HOG over the entire ROC curve, for both experimental designs. In conjunction with the results in [7], where gprHOG performed relatively poorly compared to other modern algorithms, this strongly suggests that HOG is not competitive with other modern algorithms. Given the widespread popularity of HOG, this is an important finding and suggests gprHOG (and especially HOG) should be used with caution. We note that this study uses the same FAR ranges as [7], permitting direct comparison.

## VI. CONCLUSIONS AND IMPLICATIONS FOR HOG

In this work, we considered the design of the popular HOG algorithm, as applied for GPR-based BTD by Torrione et. al [5]. We presented three modifications to HOG with the goal of better adapting it to GPR data. Using a large dataset of real-world GPR data, and two popular experimental designs, each modification was shown to improve performance. The combination of these modifications, which we name gprHOG, outperforms the original HOG algorithm by a substantial margin.

In conjunction with the results in [7], where gprHOG performed relatively poorly compared to other modern algorithms, this work strongly suggests that HOG is not competitive with other modern algorithms. Given the widespread popularity of HOG, this is an important finding and suggests gprHOG (and especially HOG) should be used with caution in future studies. Despite the relatively poor performance of gprHOG in [7], each individual modification within gprHOG is beneficial and largely generic, and therefore could be used to improve other GPR-based algorithms.

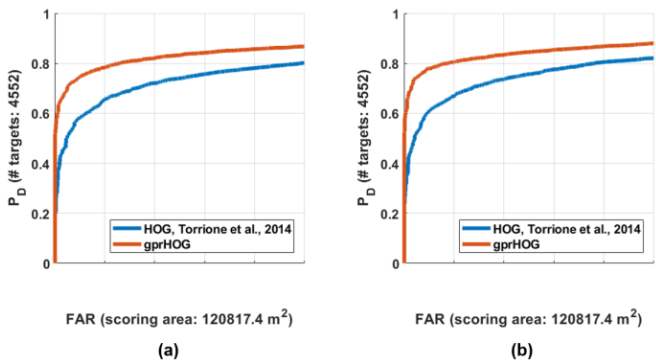

Figure 6: Performance comparison between the initial HOG algorithm as described in Torrione et al., 2014 [5], gprHOG before and after tuning to the current dataset. Performance is evaluated using (a) LBCV, (b) OBCV.

## ACKNOWLEDGEMENT

This work was supported by the U.S. Army RDECOM CERDEC Night Vision and Electronic Sensors Directorate, via a Grant Administered by the Army Research Office under Grant W911NF-13-1-0065. We would also like to thank Dr. Mark DeLong and Andy Ingham at the Duke University Office of Information Technology for technical support for this work.